n%

\documentclass[letterpaper, 10 pt, conference]{ieeeconf}  %

\IEEEoverridecommandlockouts                              %

\usepackage[textsize=scriptsize, disable]{todonotes}
\newcommand{\todomw}[1]{\todo[fancyline,color=green!40]{MW: #1}\xspace}

\usepackage{amsmath}%
\usepackage{amssymb}%
\usepackage{graphics}%
\usepackage[tight]{subfigure}
\usepackage{tabularx}
\usepackage{booktabs}
\usepackage[ruled]{algorithm2e}
\usepackage{bm}
\usepackage{algorithmic}
\usepackage{wrapfig}
\usepackage[square, numbers, sort&compress]{natbib}
\usepackage[pagebackref=true,breaklinks=true,letterpaper=true,colorlinks,bookmarks=false]{hyperref}
\usepackage{float}
\usepackage{flushend}

\usepackage{tikz}

\newcommand{\nlimb}{\textsc{N-Limb}\xspace}
\title{\LARGE \bf 
\nlimb: Neural Limb Optimization for Efficient Morphological Design}

\author{Charles Schaff and Matthew R.\ Walter%
\thanks{The authors are with the Toyota Technological Institute at Chicago,
        Chicago, IL, USA
        {\tt\small \{cbschaff,mwalter\}@ttic.edu}}%
}

\begin{document}

\maketitle
\thispagestyle{empty}
\pagestyle{empty}

\begin{abstract}

A robot's ability to complete a task is heavily dependent on its physical design.
However, identifying an optimal physical design and its corresponding control policy is inherently challenging.
The freedom to choose the number of links, their type, and how they are connected results in a combinatorial design space, and the evaluation of any design in that space requires deriving its optimal controller.
In this work, we present \nlimb, an efficient approach to optimizing the design and control of a robot over large sets of morphologies.
Central to our framework is a universal, design-conditioned control policy capable of  controlling a diverse sets of designs.
This policy greatly improves the sample efficiency of our approach by allowing the transfer of experience across designs and reducing the cost to evaluate new designs.
We train this policy to maximize expected return over a distribution of designs, which is simultaneously updated towards higher performing designs under the universal policy.
In this way, our approach converges towards a design distribution peaked around high-performing designs and a controller that is effectively fine-tuned for those designs.
We demonstrate the potential of our approach on a series of locomotion tasks across varying terrains and show the discovery novel and high-performing design-control pairs.

\end{abstract}

\section{Introduction}\label{sec:intro}

In this work, we study the efficient and automatic co-optimization of a robot's physical structure as well as the corresponding controller.
The morphological design of a robot is inherently coupled with its control policy. Traditional approaches either reason separately over design and control or perform design-control optimization by iteratively updating and fabricating 
the current set of candidate designs, optimizing their corresponding controllers, and evaluating the performance of these design-control pairs.
Data-driven approaches to co-optimization provide a promising alternative to improve the efficiency of this processes and the quality of the resulting robots.%

However, the search over discrete morphologies and their controllers is inherently challenging. The optimization process is two-fold, requiring: 1) a search over a prohibitively large, discrete, and combinatorial space of morphologies, and 2) the evaluation of each examined morphology by solving for its optimal controller.
The second step is particularly non-trivial and computationally expensive, especially for learned controllers, which makes it challenging to examine a large set of morphologies.

\begin{figure}[t]
    \centering
    \includegraphics[width=\columnwidth]{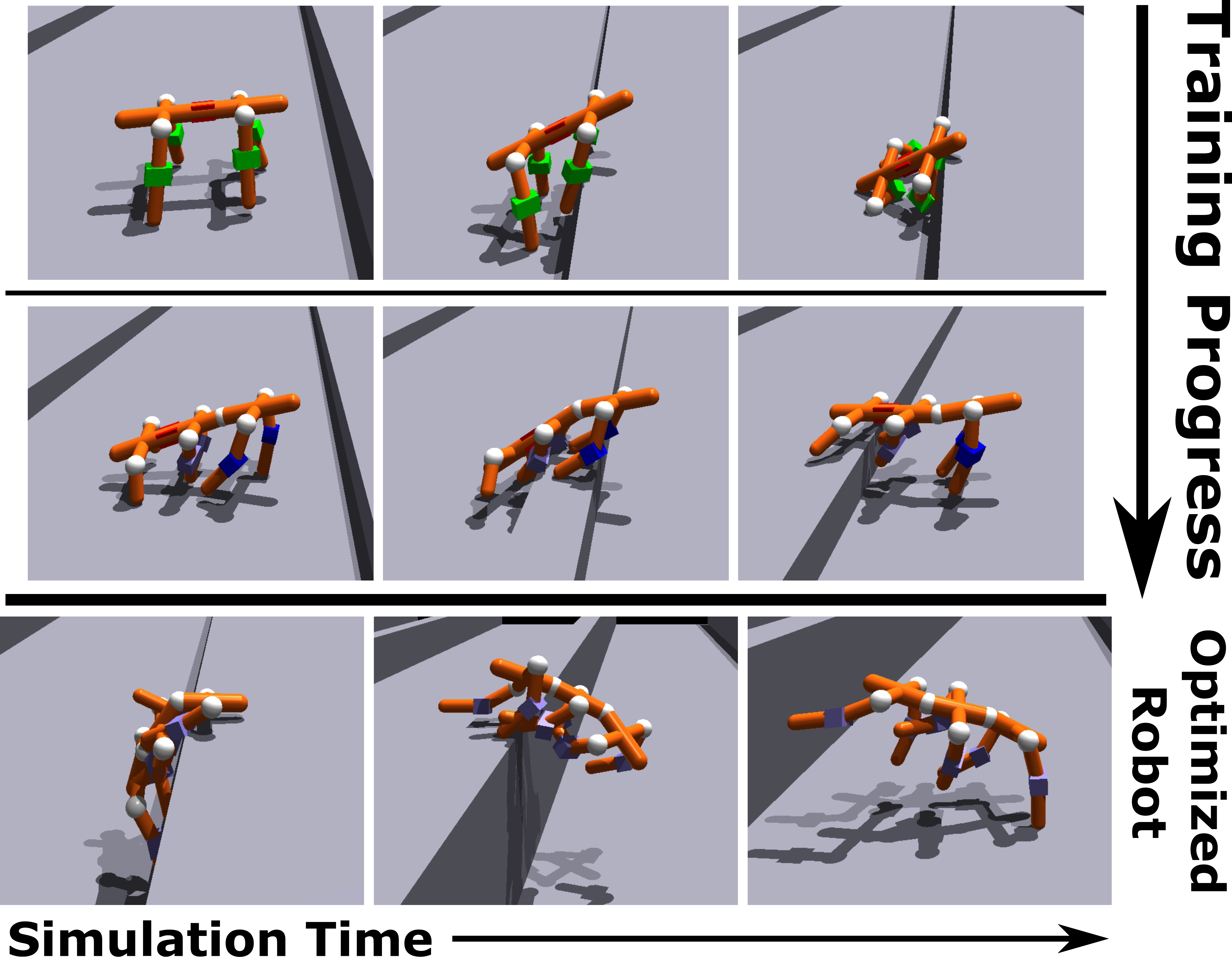}
    \caption{Our approach produces optimized robots and controllers for a given task by searching over large sets of morphologies. This figure shows the optimization of a wall-climbing robot from early in the training process (top row) to the final design (bottom row).}\label{fig:motivation}
\end{figure}

We present Neural Limb Optimization (\nlimb), an efficient algorithm for computing optimal design-control pairs across a large space of morphologies (Fig.~\ref{fig:motivation}). \nlimb is a non-trivial extension our earlier work that provides a model-free approach to co-optimization over continuous design parameters, together with a learned controller for robots with \textit{a single, fixed morphology}~\citep{schaff18}.
That approach exploits the shared structure of each design to train a universal, design-conditioned controller that is able to generalize across the space of valid designs, thus avoiding the need to solve an optimal control problem for each design.
This controller is trained in expectation over a distribution of designs, which is then subsequently updated towards higher-performing designs.
Adapting these components (i.e., a universal controller and design distribution) to a large set of morphologies is challenging.
In fact, developing universal controllers that generalize across morphologies has emerged as an important topic independent of morphological optimization~\cite{Huang2020, Kurin2021, gupta2022metamorph}. 

Our \nlimb architecture overcomes these challenges to jointly optimize robot design and control over a large space of morphologies through three key contributions.
Our approach represents the space of morphologies with a context-free graph grammar.
This has two benefits: it allows for the easy incorporation of fabrication constraints and inductive biases into the symbols and expansion rules of the grammar; and it provides a way to iteratively generate designs by sampling expansion rules.
The latter allows us to define complex, multi-modal distributions over the space of morphologies via a novel autoregressive model that recursively applies expansion rules until a complete graph has been formed.
We parameterize the universal controller with a morphologically-aware transformer architecture. We evaluate the effectiveness of \nlimb on a series of challenging locomotion tasks, demonstrating the ability to learn complex, high-performing designs coupled with agile control policies.
For videos and code, please visit our webite: \href{https://sites.google.com/ttic.edu/nlimb}{https://sites.google.com/ttic.edu/nlimb}. %

\section{Related Work} \label{sec:related_work}

\textbf{Co-optimization of robot design and control:} There exists a large body of work that addresses the problem of optimizing the physical design of a robot along with the corresponding control policy.
Initial research on co-optimization is concerned with identifying the discrete morphology %
of high-performing robots~\citep{sims1994, lipson00, murata2007self}. This morphology may take the form of a composition of fixed-size 3D blocks~\citep{sims1994} or deformable voxels~\citep{cheney2014unshackling, cheney2018scalable, spielberg2019learning}, or as a topology of rigid parts connected by fixed or actuated joints~\citep{hornby2003generative, desai2018interactive, ha2018computational, pathak2019learning, wang2019neural,zhao2020robogrammar,hejna2021task,xu2021multi}, %
which is the approach that we take here. %

A common approach to optimizing over discrete designs %
is to perform evolutionary search (ES) over a population of candidate designs~\citep{sims1994,lipson00,paul01,watson2002embodied,pfeifer2006body,murata2007self,wampler2009optimal,hiller2011automatic,bongard11,bongard2013evolutionary,cheney2014unshackling,brodbeck2015morphological,cheney2018scalable,pan2021emergent}. %
However, ES is prone to local minima and is sample-inefficient, in part due to the need to maintain separate control policies for each candidate design in the population. Our approach improves upon the sample-efficiency of these methods by sharing experience across designs through a single control policy.
Alternatively, several approaches improve sample efficiency by leveraging additional knowledge or assumptions about the system dynamics, typically in the context of optimizing the continuous parameters of a fixed morphology.
These approaches include trajectory optimization~\citep{spielberg17, bravo2020one}, linear approximations to the dynamics~\citep{ha17}, and leveraging differentiable simulators~\citep{xu2021end}.

Similar to our approach, several reinforcement learning-based strategies to co-optimization exist for both continuous~\cite{schaff18, ha2018reinforcement, luck_data-efficient_2019, chen2020hardware} and discrete design spaces~\citep{spielberg2019learning, schaff2022soft, pathak2019learning}. 
\citet{luck_data-efficient_2019} use a soft actor-critic algorithm and use a design-conditioned Q-function to evaluate designs. 
\citet{chen2020hardware} model the design space as a differentiable computational graph, which allows them to use standard gradient-based methods.  
In the context of discrete design spaces, \citet{spielberg2019learning} propose an autoencoder-based method that jointly optimizes the placement and control of a large number deformable voxels for soft-body locomotion tasks.
\citet{yuan2021transform2act} represent the design generation process as part of the environment and train a dual-purpose design generation and control policy.

\textbf{Universal control policies:} Traditional population-based co-optimization strategies are difficult to scale due to the need to maintain separate controllers for each design within the population as it changes over time.
Recent approaches adopt a single design-aware control policy to improve optimization efficiency. \citet{schaff18} and \citet{won2019learning} propose approaches for optimization within fixed morphology. %
More sophisticated controllers leverage graph neural networks (GNN)~\citep{scarselli2008graph} structured according to the robot morphology~\citep{wang2018nervenet, pathak2019learning, wang2019neural, Huang2020, yuan2021transform2act}.
Recently, shared controllers based on self-attention mechanisms~\citep{Kurin2021, gupta2022metamorph} have outperformed GNNs due to the ability to propagate information across nodes more effectively than the message-passing schemes of GNNs.
Motivated by the empirical success of self-attention over GNNs, we leverage transformers~\citep{vaswani2017attention} to model our controller.
\section{Co-optimization of Design and Control}

In this section, we introduce the problem of jointly optimizing the physical design and corresponding control of a robot in the context of a specified task.

\subsection{Problem Definition}

\newcommand{\mS}{\mathcal{S}}
\newcommand{\mA}{\mathcal{A}}
\newcommand{\mP}{\mathcal{P}}
\newcommand{\mR}{\mathcal{R}}
\newcommand{\mE}{\mathcal{E}}
\newcommand{\mM}{\mathcal{M}}

We formulate the problem of co-optimizing design and control as a set of related reinforcement learning problems.
Given a set of robot designs $\Omega$ and a task definition, we define a \textit{design-specific} Markov decision process (MDP) for each $\omega \in \Omega$: $\mM_\omega = \textrm{MDP}(\mS_\omega,
\mA_\omega, \mP_\omega, \mR_\omega)$, where $\mS_\omega$ is the state space, $\mA_\omega$ is the action space, $\mP_\omega: \mS_\omega \times \mA_\omega \times \mS_\omega \rightarrow [0,1]$ is the transition dynamics, and $\mR_\omega: \mS_\omega \times \mA_\omega \times \mS_\omega \rightarrow \mathbb{R}$ is the reward function.
It is natural that these MPDs share some common structure, e.g., each $\mR_\omega$ encodes the same objective, and the dynamics and the state and action spaces differ only with respect to changes in $\omega$.

Let $\pi^*_\omega: \mS_\omega \times \mA_\omega \rightarrow [0, 1]$ be the optimal policy for MDP $\mM_\omega$ and $\mathbb{E}_{\pi^*_\omega}\left[\sum_t \gamma^t r_t\right]$ be its expected return. The goal of co-optimization is to find the optimal design-controller pair $(\omega^*$, $\pi^*_{\omega^*})$ such that:
\begin{equation}
    \label{eqn:objective}
    \omega^*, \pi^*_{\omega^*}  = \underset{\omega, \pi^*_\omega}{\text{arg max}} \; \mathbb{E}_{\pi^*_\omega}\left[\sum_t \gamma^t r_t \right].
\end{equation}

\subsection{Co-optimization via Universal and Transferable Policies}

\begin{figure*}[t!]
    \centering
    \includegraphics[width=\textwidth]{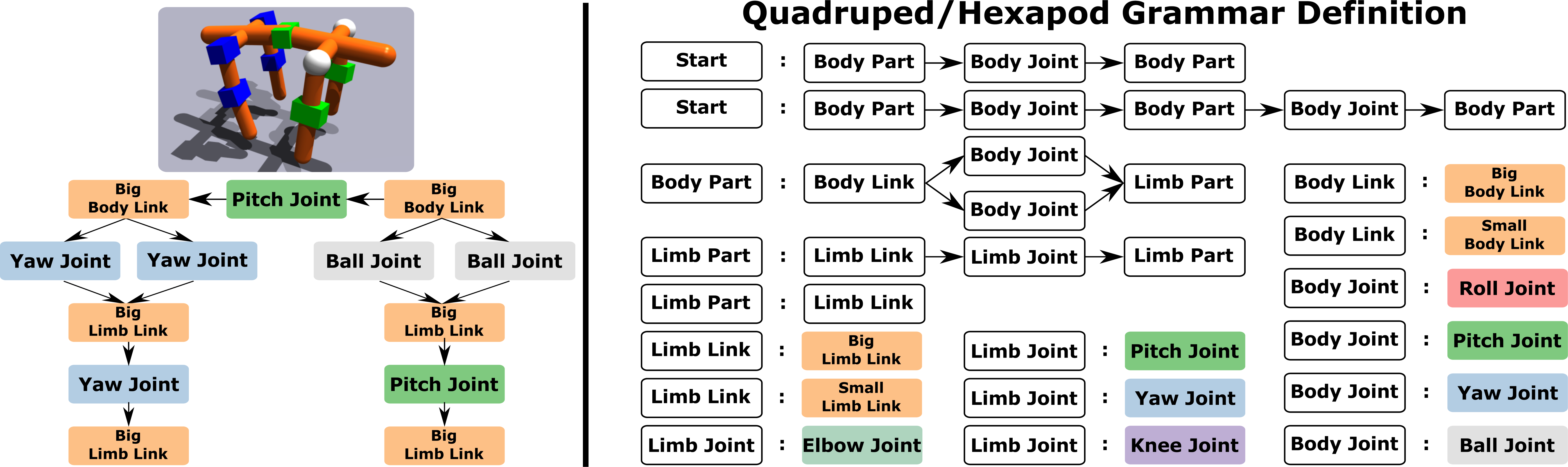}
    \caption{Our approach defines the space of valid designs using a context-free graph grammar in which terminal symbols denote robot parts and expansion rules describe how those parts can be combined. \textbf{Left}: An example design and its corresponding graph structure. \textbf{Right}: The definition of a bio-inspired grammar that produces quadruped and hexapod designs. Nonterminal symbols are denoted with white boxes, terminal symbols with colored boxes, and expansion rules are denoted with a colon. %
    }\label{fig:grammar}
\end{figure*}

Based on Equation \ref{eqn:objective}, the search over the design space $\Omega$ requires access to the optimal controller $\pi^*_\omega$ for each design.
Obtaining these controllers can be challenging, as it is often computationally intractable to train independent controllers for each design.
However, we can leverage the shared structure between each design to efficiently train a \textit{design-conditioned} controller: $\pi: \mS \times \mA \times \Omega \rightarrow [0, 1]$.
Ideally, this controller would serve as a proxy for optimal controllers as well as generalize zero-shot to unseen designs in the design space.
Implementing such a controller has been studied in the context of co-optimization~\citep{schaff18, yuan2021transform2act, chen_hardware_2020} as well as the transfer of control policies across morphologies~\citep{Huang2020, Kurin2021, gupta2022metamorph}.

With such a control policy, the co-optimization procedure reduces to a search over the design space $\Omega$.
In this work, we use a zero-order search algorithm based on policy gradients that optimizes a distribution over the design space towards higher performing designs.
Let $p_\phi$ be a distribution over $\Omega$ parameterized by $\phi$ and $\pi_\theta$ be a design-conditioned policy.
Then, our training objective is to maximize the expected return of $\pi_\theta$ under the design distribution $p_\phi$:
\begin{equation}
    \label{eqn:train}
    \phi^*, \theta^*  = \underset{\phi, \theta}{\text{arg max}} \; \mathbb{E}_{\omega \sim p_\phi}\mathbb{E}_{\pi_\theta}\left[\sum_t \gamma^t r_t \right].
\end{equation}

The controller can then be trained using any standard RL algorithm and the design distribution can be updated with any zero-order method. Similar to past work~\citep{schaff18}, we use policy gradients to update both the policy and the design distribution.
This leads to the following update equations:
\begin{align}
    \label{eqn:update}
    \nabla_\phi &= \mathbb{E}_{\omega \sim p_\phi}\left[\nabla_\phi\text{log }p_\phi(\omega)\mathbb{E}_{\pi_\theta}\left[\mR_t \right] \right] \nonumber \\
    \nabla_\theta &= \mathbb{E}_{\omega \sim p_\phi}\left[\mathbb{E}_{\pi_\theta}\left[\nabla_\theta \text{log }\pi_\theta(s_t, a_t)\mR_t\right] \right],
\end{align}
where $\mR$ is the expected return at time $t$.
This algorithm trains a controller to maximize performance in expectation over the design distribution $p_\phi$.
When this controller is sufficiently trained, it is used as a proxy for the optimal controllers of each design and $p_\phi$ is updated towards higher performing designs. Then training continues until $p_\phi$ converges on a single design and $\pi_\theta$ is optimized for that design.
See Algorithm~\ref{alg:main} for details.

\begin{algorithm}[t] \label{alg:main}
    \caption{Joint Optimization of Design and Control} \label{algorithm}
    \begin{algorithmic}[1]
        \STATE Initialize $\pi_\theta(a \vert s,\omega)$, $p_\phi(\omega)$, $T=0$, $T_0$
        \WHILE{$T < \textrm{BUDGET}$}
            \STATE Sample designs $\omega_1, \omega_2, \hdots, \omega_n \sim p_\phi$ \label{alg:sample}
            \STATE Control $\omega_1, \omega_2, \hdots, \omega_n$ with $\pi_\theta$ for $t$ timesteps \label{alg:control}
            \STATE Update $\theta$ using \textrm{PPO} with collected trajectories \label{alg:ppo}
            \STATE Set timestep $T = T + nt$
            \IF {$T > T_0$}
                \STATE Compute average episode returns $R_{\omega_1}, R_{\omega_2}, \hdots, R_{\omega_n}$
                \STATE Update $\phi$ using $\nabla_\phi \approx \sum_{i=0}^n \nabla \text{log }p_\phi(\omega_i) R_{\omega_i}$
            \ENDIF
        \ENDWHILE
    \end{algorithmic}
\end{algorithm}
%
\section{\nlimb: Optimizing Design and Control over Morphologies}
\label{sec:nlimb}

While \citet{schaff18} successfully employ the above approach to find optimal design-control pairs, their approach is limited to simple, pre-defined kinematic structures and is not able to optimize across morphologies. We propose a framework that builds on this approach to perform morphological optimization. 

\begin{figure*}[t!]
    \centering
    \subfigure[Actor-critic architecture]{\includegraphics[width=\textwidth]{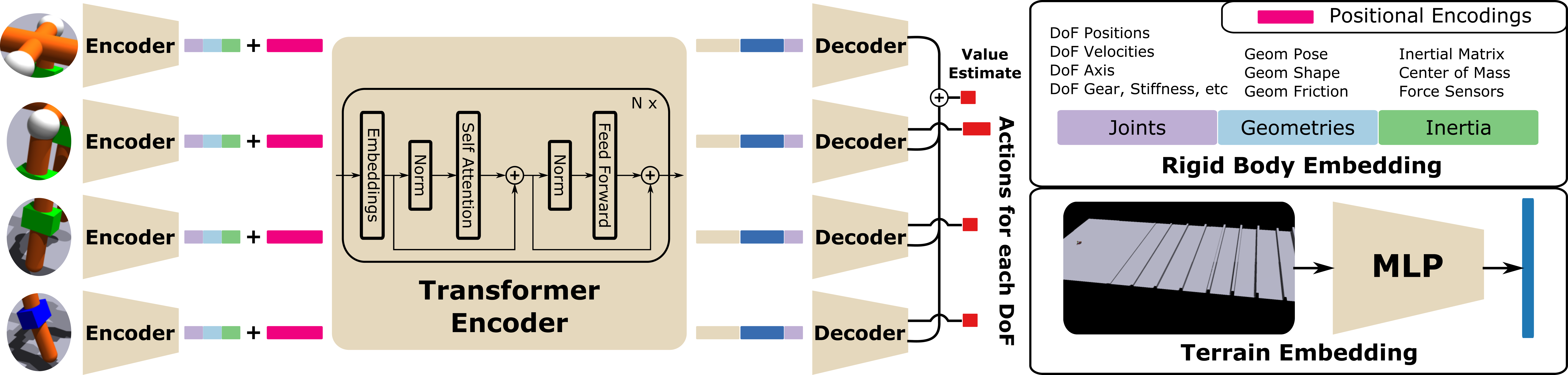}\label{fig:policy}}\\
    \subfigure[Autoregressive model for design generation]{\includegraphics[width=\textwidth]{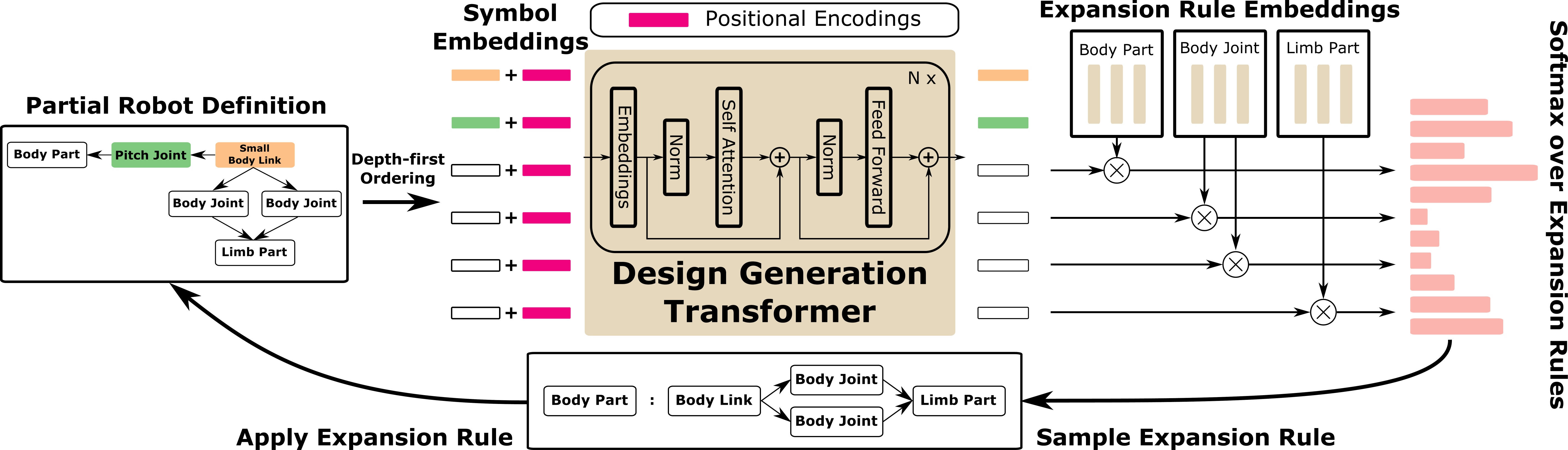}\label{fig:design-generation}}
    \caption{\textbf{(a)} Self-attention based actor-critic architecture. Rigid body pose and attributes are encoded and processed via a transformer encoder to produce actions for each DoF and a value estimate. \textbf{(b)} The design distribution is modeled via an autoregressive transformer architecture. Partial graphs are processed by the model to produce a distribution over expansion rules of the grammar. Rules are sampled and applied until a complete graph is formed.}
\end{figure*}

\subsection{Graph Grammars for Defining Design Spaces across Morphologies}

In this work, we explore the use of context-free graph grammars to define the space of valid morphologies.
We define a context-free graph grammar as the tuple $(N, T, R, S)$, where $N$ denotes the set of non-terminal symbols, $T$ denotes the set of terminal symbols, $R = \{r_s^i = (s, G_{s}^i)\}$ is the set of production rules that map a non-terminal symbol $s \in N$ to a graph $G_s^i$, and $S$ is the starting graph. The application of a production rule replaces a node with non-terminal symbol $s$ with the corresponding graph $G_s^i$. Nodes within these graphs express limb information (e.g., geometry, pose, mass, inertia, etc.), and edges model joint information (e.g., type, kinematic transformation between limbs, max torque, gear ratio, etc.).
Additionally, using a grammar allows the user to tailor the search towards plausible designs by incorporating fabrication constraints or structural biases (e.g., left-right symmetry).

In our experiments, we use the grammar depicted in Figure~\ref{fig:grammar} to generate quadruped and hexapod robots as a composition of various joint types and limbs with different numbers of parts and sizes, maintaining left-right symmetry. The resulting space includes more than 2.5M designs.

\subsection{Universal Controllers via Design-aware Transformers}
Our co-optimization algorithm requires a universal controller that can act as a proxy for the optimal controller across the design space $\Omega$. 
Control policies that employ graph neural networks~\citep{wang2018nervenet,Huang2020} as well as those that use self-attention~\cite{Kurin2021, gupta2022metamorph} have proven effective at controlling a variety of morphologies.
In this work, we use the transformer architecture~\cite{vaswani2017attention} to parameterize an actor-critic network as it was shown to outperform graph neural networks~\cite{Kurin2021}.
For an overview of our actor-critic architecture, see Figure~\ref{fig:policy}.

At each point in time, the agent receives various types of state information: local information about each rigid body and joint (e.g., pose, velocity, etc.), morphological information (e.g., shape, inertial, joint axes, etc.), and global sensory information (e.g., local terrain observations). Our architecture consists of three core components: the first encodes local and morphological information; the second processes that information through self-attention and then appends global information; and the third module decodes actions and a value estimate.

\textbf{Rigid-body embeddings:} We encode each robot by flattening its graph structure into a sequence of rigid-body embeddings using a depth-first traversal. Each of these embeddings contains local pose and morphological information about the rigid body, as well as each controllable degree-of-freedom that connects it to its parent. 
Information about geometries, degrees-of-freedom, and inertia are encoded separately using two-layer MLPs and concatenated to obtain the rigid-body embedding.

\begin{figure*}[!t]
    \centering
    \includegraphics[width=\linewidth]{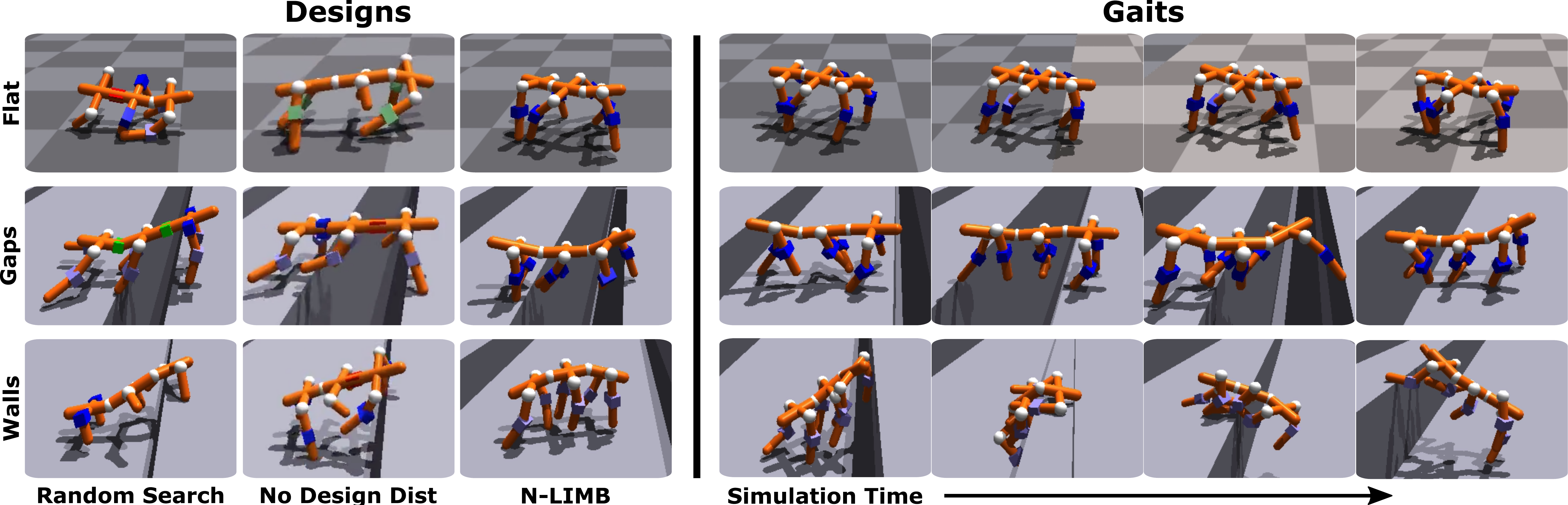}
    \caption{\textbf{Left:} A comparison between the designs learned by \nlimb, those of random search and the ablation of the design distribution for the three terrain types. \textbf{Right:} A visualization of the learned gaits for the designs found by \nlimb.}
    \label{fig:designs-gaits}
\end{figure*}

\textbf{Process:}
The rigid-body embeddings are combined with positional encodings and processed with a transformer encoder.
Similar to \citet{gupta2022metamorph}, we postpone the fusion of global information, such as observations about the surrounding terrain, until after the self-attention layers to both reduce the number of transformer parameters as well as avoid diluting the low-dimensional pose and morphological information. 
Terrain observations are comprised of sparsely sampled height-field information that is encoded into a flat embedding vector using a three-layer MLP and concatenated with each processed rigid-body embedding. 

\textbf{Decoding actions and value estimates:}
Our model decodes these representations to produce a value estimate by averaging the outputs of a value-decoder MLP across each rigid-body.
We concatenate an embedding of each degree-of-freedom associated with a rigid body and apply an action-decoder MLP that produces the mean and standard deviation of a Gaussian distribution.

\subsection{Autoregressive Models for Design Generation}

Given the definition of a graph grammar, designs can be generated by autoregressively sampling expansion rules until the resulting graph contains only terminal symbols.
To build such an autoregressive model, we again use the transformer architecture~\cite{vaswani2017attention}.
Sampling designs in this way allows, in principle, for arbitrary distributions over the design space that can capture multi-modal behavior and complex dependencies between symbols.
For an overview of our model, see Figure~\ref{fig:design-generation}.

At each stage of generation, the model receives as input a partial graph $G$.
This graph is then flattened to a sequence of symbols $s_1$, \dots, $s_n$ using a depth-first traversal. Those symbols are then embedded using an embedding table, combined with a positional encoding, and processed with a transformer encoder to produce representations $h_{s_1}, \ldots, h_{s_n}$. 
We decode logits for each expansion rule by computing a dot-product between each representation $h_{s_i}$ and learnable embeddings corresponding to each expansion rule $r_{s_i}^j$ associated with symbol $s_i$.
These logits are combined across all symbols in $G$ using a softmax operation.
An expansion rule is then sampled from this distribution, used to update the graph $G$, and then this process is repeated until $G$ contains only terminal symbols. %

\section{Results} \label{sec:experiments}

\begin{figure*}[!t]
    \centering
    \includegraphics[width=\linewidth]{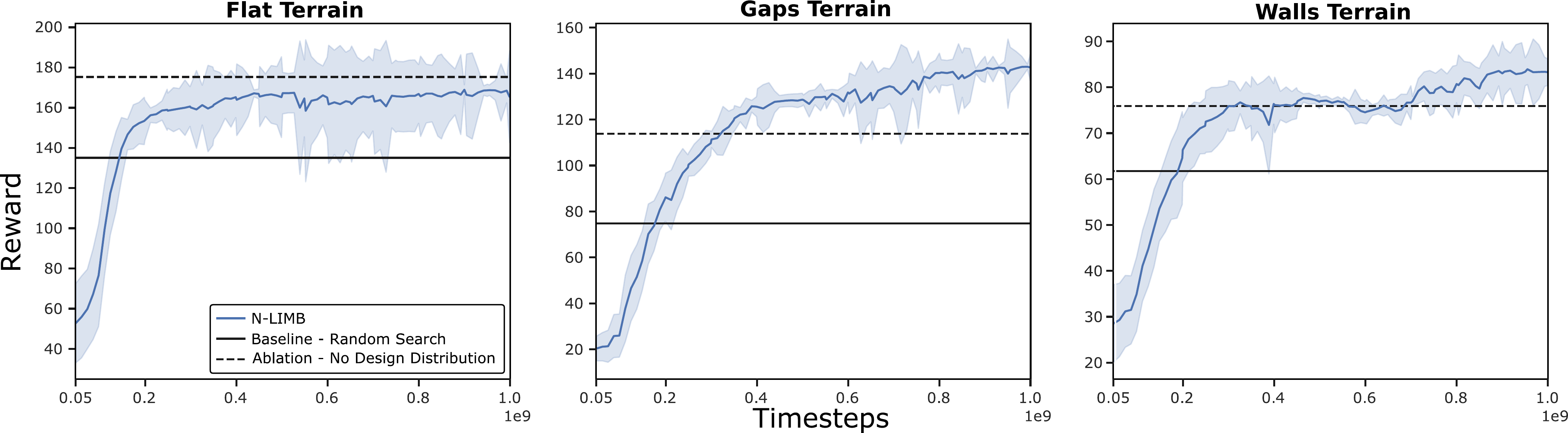}
    \caption{Plots that show the reward (mean and standard deviation) of designs sampled from the design distribution over the course of training for the three terrain types. Dashed horizontal lines denote the rewards of the highest performing design-control pairs for the baseline and ablation algorithms.} \label{fig:reward-evolution}
\end{figure*}
We evaluate our \nlimb framework by performing co-optimization of design and control for locomotion tasks over a variety of terrains.
In all cases, \nlimb efficiently finds high-performing design-control pairs that outperform those of competitive baselines.

\subsection{Experiment Details}
We optimize and evaluate design-control pairs using the IsaacGym GPU-based physics simulator~\cite{makoviychuk2021isaac}, allowing us to train with thousands of parallel environments on a single GPU.
The reward function for the locomotion task is comprised of rewards for forward progress, staying upright, maintaining forward heading, and penalties for large torques, energy cost, and joints reaching their limits. %
Episodes are terminated early if the robot has fallen over.
We consider three different terrain types: (i) flat terrain (``Flat''); (ii) a terrain with randomly placed gaps of random widths that the robot must cross (``Gaps''); and (iii) a terrain with randomly placed walls of different heights that the robot must climb over (``Walls'').
We run the \nlimb algorithm on a single NVIDIA A4000 GPU for $1$B timesteps (roughly two days).
We use Proximal Policy Optimization~\cite{schulman17} to train both the controller and design distribution.\todomw{Remvoved the reference to the Appendix since ICRA doesn't allow us to submit supplementary material. We should add back for arXiv} %

\subsection{Baselines and Ablations}
We compare against one baseline approach and one ablation.
To ensure a fair comparison with our approach, we provide both with the same computational budget as \nlimb. 
Similar to previous work in co-optimization~\citep{schaff18,zhao2020robogrammar,hejna2021task}, we compare our framework against a decoupled approach that randomly samples designs from the hexapod grammar and trains controllers for each design individually, choosing the resulting design-control that yields the best performance. We refer to this approach as \textit{Random Search}.
Next, we compare against an ablation that involves removing the the design distribution from our algorithm. The ablation trains a universal controller as is done for \nlimb, and then performs random search using this controller to evaluate designs. We then individually train controllers for the top designs until the computational budget is reached and select the best-performing design-control pair. This allows us to gauge the contributions of the universal controller and the autoregressive design distribution separately.

\subsection{Results}
Across all three terrains, we find that \nlimb is able to outperform the random search baseline by a large margin.
Figure~\ref{fig:designs-gaits} shows the best designs found by \nlimb and the baselines, as well as the learned gaits while traversing difficult sections of each of the three terrain types. While \nlimb initially favors quadruped designs in some cases (Fig.~\ref{fig:motivation}, top row), it converges to hexapod designs for all three terrain types. \nlimb chooses to connect the three body links with ball joints, suggesting that the benefits of having additional degrees-of-freedom outweigh a reduction in available torque and narrower joint limits.
The choice of a hexapod design may be a result of the fact that achieving longer bodies with our grammar requires a design with six legs.
Indeed, on the Gaps terrain (Fig.~\ref{fig:designs-gaits}, middle row), \nlimb selects a design that has the largest body available within the grammar through the use of larger body links, which is necessary to cross wider gaps. It identifies leg links that are connected by ``yaw'' joints.
On the Walls terrain, however, the optimal design (Fig.~\ref{fig:designs-gaits}, bottom row) has a smaller body and connects limb links with ``knee'' joints, which can fold back $180^\circ$, allowing the robot to fold its limbs while cresting or pushing off the wall.
For locomotion on the Flat terrain, \nlimb converges to a design identical to that of the Gaps terrain, except that the torso is comprised of three small (vs.\ large) body links, trading off the need to span wide gaps for lower-torque bounding gates.

Figure~\ref{fig:reward-evolution} shows the quantitative performance of \nlimb compared against the baseline. 
It displays the mean and standard deviation of returns obtained by designs sampled from the design distribution throughout training. Early on, the ability for the policy to control the set of morphologies  drawn from the design distribution is limited. However, as the controller improves and \nlimb updates the design distribution, we see that the algorithm yields designs paired with the universal control policy that quickly outperform the baseline.
Additionally, ablating the design distribution leads to reduced performance on the Gaps and Walls terrains, but improves significantly over the random search baseline. This shows that our approach benefits both from the improved efficiency of the universal controller and the iterative refinement of potential designs.

\begin{figure}[!t]
    \centering
    \includegraphics[width=\linewidth]{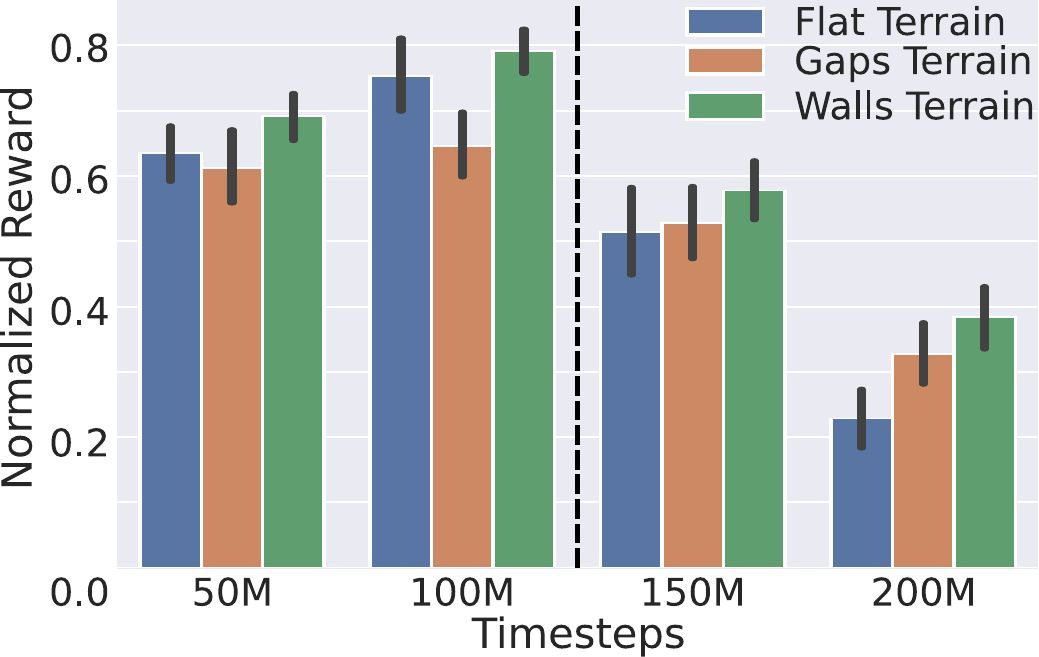}
    \caption{A visualization of the generalizability of the universal \nlimb controller as a fraction of the reward achieved using controllers trained separately on designs sampled from the initial design distribution. In all three domains, the relative reward improves until, after $100$M timesteps (dashed line), we begin to update the design distribution, upon which the \nlimb policy learns to specialize to higher-performing designs.}\label{fig:generalization}
\end{figure}
\subsection{Generalization of the Universal Controller}
Crucial to our approach is a universal controller that provides a proxy for the optimal controller when evaluating a design.
If this controller is unable to provide a reasonable estimate of a design's optimal performance, the design distribution may incorrectly focus the search around suboptimal designs.
To verify that our controller provides a reasonable proxy, we compare it against controllers trained individually on each design sampled from the initial design distribution.
Figure~\ref{fig:generalization} shows the performance of the universal controller as a fraction of the reward of the individually trained controllers at four checkpoints during training.
We keep the design distribution fixed for the first $100$M steps of training, during which the universal controller impressively obtains up to $80$\% of the performance of individually trained controllers.
Once we begin updating the design distribution (after $100$M timesteps), we see an expected decline in performance as the controller specializes to the shifting design distribution.

\section{Conclusion}\label{sec:conclusion}

We presented \nlimb, an efficient and effective approach to co-optimizing robots and their controllers across large sets of morphologies.
\nlimb formulates the set of valid morphologies as a context-free graph grammar, allowing users to focus the search on realizable robots by incorporate fabrication constraints and inductive biases.
Given such a grammar, we parameterize a distribution over the design space using a novel autoregressive model that samples the expansion rules of the grammar until a completed robot is formed.
The optimization is carried out by training a universal controller in expectation over the design distribution, while simultaneously shifting that distribution towards higher performing designs.
In this way, the optimization converges to a design-control pair that is optimal for the given task.
We demonstrate the potential of our approach by learning high-performing design-control pairs on a variety of locomotion tasks and terrains. Future work will investigate the real-world performance of the resulting designs, building off of our recent work on sim-to-real transfer for co-optimization~\cite{schaff2022soft} and successful approaches to transfer for control~\citep{tan18, zhao2020sim}.

 \flushcolsend

\clearpage

\bibliography{refs}  %
\flushcolsend

\end{document}